\documentclass[10pt,twocolumn,letterpaper]{article}

\usepackage{cvpr}
\usepackage{times}
\usepackage{epsfig}
\usepackage{graphicx}
\usepackage{amsmath}
\usepackage{amssymb}
\usepackage{bbm}
\usepackage{mathtools}
\usepackage{multirow}
\usepackage[ruled,vlined]{algorithm2e}

\usepackage{caption}
\def\mL{{\mathcal L}}

\usepackage[pagebackref=true,breaklinks=true,letterpaper=true,colorlinks,bookmarks=false]{hyperref}

\cvprfinalcopy % *** Uncomment this line for the final submission

\def\cvprPaperID{7057} % *** Enter the CVPR Paper ID here

\ifcvprfinal\fi
\begin{document}

%%%%%%%%% TITLE
\title{Switchable Precision Neural Networks}

\author{Luis Guerra$^1$
\and
Bohan Zhuang$^2$
\and
Ian Reid$^2$
\and
Tom Drummond$^1$~~~~~~~~
\and
$^1$Monash University
\and
$^2$The University of Adelaide
}

\affiliation{}

\maketitle
%\thispagestyle{empty}

%%%%%%%%% ABSTRACT
\begin{abstract}
   Instantaneous and on demand accuracy-efficiency trade-off has been recently explored in the context of neural networks slimming. 
   %In this paper we leverage switchable batch normalization to propose a complementary strategy to add an additional degree of flexibility by training a shared network capable of operating at multiple quantization levels.
   %Quantizing both weights and activations allows for different memory, power consumption and inference time configurations.
   In this paper, we propose a flexible quantization strategy, termed Switchable Precision neural Networks (SP-Nets), to train a shared network capable of operating at multiple quantization levels. At runtime, the network can adjust its precision on the fly according to instant memory, latency, power consumption and accuracy demands.
   For example, by constraining the network weights to 1-bit with switchable precision activations, our shared network spans from BinaryConnect to Binarized Neural Network, allowing to perform dot-products using only summations or bit operations. In addition, a self-distillation scheme is proposed to increase the performance of the quantized switches. We tested our approach with three different quantizers and demonstrate the performance of %quantizable models 
   SP-Nets against independently trained quantized models in classification accuracy for Tiny ImageNet and ImageNet datasets using ResNet-18 and MobileNet architectures.
\end{abstract}

\section{Introduction}

Deep Neural Networks (DNNs) have achieved great success in a wide range of vision tasks, such as image classification \cite{he2016deep}, semantic segmentation \cite{long2015fully} and object detection \cite{redmon2016you}.
%however the gap between flexible, resource unconstrained experiments and real world practical applications is still large. 
However, the large model size and expensive computational complexity remain great obstacles for many applications, especially on some constrained devices with
limited memory and computing resources.
%The reason is partially due to hardware constraints, specifically memory footprint, power consumption and inference time. 
Network quantization is an active field of research focusing on alleviating such issues. In particular, \cite{courbariaux2015binaryconnect, hubara2016binarized, rastegari2016xnor} set the foundations for 1-bit quantization, while \cite{hubara2016quantized, zhou2016dorefa} for arbitrary bitwidth quantization. Progressive quantization \cite{bai2019proxquant, ajanthan2019proximal, zhuang2018towards, sakr2018true}, loss aware-quantization \cite{hou2018loss, zhou2018explicit}, improved gradient estimators for non-differentiable functions \cite{liu2018bi} and RL-aided training \cite{liu2019training}, have focused on improved training schemes, while mixed precision quantization \cite{tung2018clip}, hardware-aware quantization \cite{wang2018haq} and architecture search for quantized models \cite{shen2019searching} have focused on alternatives for standard quantized models. However, these strategies are exclusively focused on improving the performance and efficiency of static networks.

Dynamic routing networks try to provide improvements using an alternative approach. By performing computations conditioned on the inputs, the networks are capable of saving resources by executing just the sufficient amount of operations required to map the input to the desired output. Popular strategies include skipping convolutional layers \cite{wang2018skipnet, campos2017skip, wu2018blockdrop} based on the input data complexity and early classifiers \cite{mcgill2017deciding}.

Our proposed approach taken by slimmable neural networks \cite{yu2018slimmable} falls in the category of dynamic networks, but following a different principle, aiming to provide on demand trade-offs, rather than input dependant. 
To the best of our knowledge, dynamic quantization of DNNs has not been explored in the literature. 
Slimmable networks provide width (number of channels) \textit{switches} that allow to perform inference utilizing only sections of the network according to on-device demands and resource constraints.
Similar to slimmable networks, we attempt to provide the first approach by developing a network whose weights and activations can be quantized at various precision at runtime, permitting instant and adaptive accuracy-efficiency trade-off, which is termed \textit{switchable precision}.
In particular, 1-bit DNNs are both an interesting and challenging case of our SP-Net. With binary weights, we can train a shared network ranging from BinaryConnect \cite{courbariaux2015binaryconnect} and Binarized Neural Network \cite{hubara2016binarized}, where the inner product can be efficiently implemented using summation or bit operations according to on-device resource constraints.
%We name our precision switchable one, providing bitwidth switches, as \textit{switchable precision}.
Furthermore, our PS-Nets frequently yields higher accuracy than
individually trained quantized networks.

%all the switches will update the accumulated running mean and variance, therefore these estimates will be erroneous. 
%Due to the inconsistency of feature mean and variance across different switches, 

%As a result, the network will be able to train without S-BN, but will incur in inaccurate test inference. Batch-norm layers additionally count with two learnable parameters, $\beta$ and $\gamma$, which are similarly updated by all the switches. In contrast, providing with per switch independent versions of $\beta$ and $\gamma$ is not as crucial, however, they provide a slight accuracy boost.
%\bohan{Note that we should explain S-BN under our quantization scenario.}

%regular quantized networks capable of operating at a single quantization level.
However, different precision switches are difficult to optimize as a whole. And we summarize the reasons in two parts.
On one hand, during training, batch normalization (BN) layers use the current batch statistic to perform intermediate feature maps normalization, while estimating the global statistics by accumulating a running mean and running variance, which are used as the replacement during testing. 
However, the batch statistics of each precision switch are different. As a result, the discrepancy of feature mean and variance across different switches leads to inaccurate accumulated statistics
of BN layers.  To solve this problem, we follow \cite{yu2018slimmable} to train a switchable precision network by using independent BN parameters for each switch, named \textit{switchable batch normalization (S-BN)}. 
%\bohan{any modification to S-BN for quantization? or just use it? No modifications, just use it}

%Universally slimmable networks (US-net) \cite{yu2019universally} extended the work of \cite{yu2018slimmable} by provinding a training scheme that allows to efficiently compute the S-BN parameters for all the avaialble width switches, provinding a continously slimmable network. In this work, we only work with only relatively small sets of quantization switches as proof of concept.
%\bohan{confusing. you use S-BN in US-Net? or just in slimmable neural networks. And these two paragraphs should be merged.}
%\luis{US-Net uses S-BN but with A Lot of switches (that's why it is continuosly slimmable). I just added it to future work}

On the other hand, we conjecture that simultaneously optimizing multiple quantized switches will reflect in the loss manifold by progressively quantizing the loss surface, and the higher precision switches will assist the lower precision ones to achieve a less noisy and smoother convergence, potentially leading them to a better minima. Conversely, the network will converge to only minimas which perform well at all the bitwidths potentially harming the performance of some of the switches, particularly the higher precision ones.
During training of SP-Nets, the gradients of each switch are combined before running an optimizer step. However, there is no explicit mechanism impeding the individual switches from moving in distinct directions. In order to encourage the different switches to move in approximately the same direction, we propose a self-distillation strategy, where the full precision switch provides a guiding signal for the rest of the switches.
Specifically, only the teacher full-precision switch sees the ground-truth while the student low-precision switches  are trained by distilling knowledge from the full-precision teacher.

%Finally, we execute multiple tests and ablation studies to showcase the performance of out approach in multiple scenarios
In order to increase the flexibility of our model, in addition to equipping the network with switchable precision representations, we extend our approach to 3 different quantizers and equip them with slimming (width switchable) capability.

%dot-products can be computed using uniquely summations. 1 bit per activation further simplifies computation by replacing dot-products with \texttt{xnor-popcount} operations. 
%Nevertheless, as explained in section \ref{sec:tanh_quantizer} special consideration must be taken into account.

Our contributions are summarized as follows:
\begin{itemize}
     \itemsep -0.125cm
    \item We leverage S-BN to train a shared network executable at different bitwidths on weights and activations according to runtime demands. 
    \item We propose a self-distillation scheme to jointly train the full-precision teacher switch and the low-precision student switches. By doing so, the full-precision switch provides a guiding signal to significantly improve the performance of the low-precision switches.
%    \item We increase the accuracy-efficiency trade-off flexibility by training a mixed slimmable/quantizable model.
    \item We investigate the effectiveness of our method on uniform and non-uniform quantization with various quantizers through extensive experiments on image classification.
    %on networks with and without residual connections.
    %Three non-linear quantizers are tested: a tanh-based quantizer on both activations and weights proposed by \cite{zhou2016dorefa}, a Rectifier Linear Unit (ReLU) approximation quantizer only on activations proposed by \cite{zhou2016dorefa, cai2017deep} and a logaritmic quantizer on weights and activations \cite{miyashita2016convolutional}.

\end{itemize}

\section{Related Work}

\noindent\textbf{Network slimming.} Slimmable networks introduced by \cite{yu2018slimmable} developed a procedure useful for training a DNN with switchable widths. The motivation is to provide instant and on demand accuracy-efficiency trade-offs. It was further generalized by \cite{yu2019universally} allowing to efficiently train a network continuously slimmable, executable at any arbitrary width. Moreover, non-uniform slimming was introduced, allowing for layer-wise width selection. The training principle behind network slimming has found applications in the fields of pruning \cite{yu2019network}, network distillation \cite{zhang2019your}, architecture search \cite{cai2019once}, adaptive inference \cite{ruiz2019adaptative} and finally, in our work, network quantization.

%DNNs quantization can be split in two major branches. Progressive quantization and quantized networks trained with Straight Through Estimator (STE) \cite{bengio2013estimating}.

%Progressive quantization has been extensively explored in two settings. First by some penalization or drive on a initial full precision model during training encouraging it to be quantized with increasing penalization hyperparameter. This setting encompasses multiple variants like \cite{hou2018loss, zhou2018explicit, bai2019proxquant, ajanthan2019proximal, sakr2018true, choi2018learning} and has recently acquired more attention given that it regularly attains higher accuracy than STE-based, due to the noisy nature of STE. Second \cite{zhuang2018towards} by starting with a full precision pre-trained model, subsequentily quantizing the model weights and activations to half-precision and retraining estimating the gradients via STE. This cycle is repeated until reaching the desired quantization levels.

%Similarly to the second setting, our method lies in the intersection of progressive quantization and STE-based quantization, however our real version of the network along with the intermediate switches are also useful.
\noindent\textbf{Network quantization.} Quantization based methods represent the network weights and/or activations with very low precision, thus yielding highly compact DNN models compared to their floating-point counterparts with considerable memory and computation savings.
BNNs~\cite{hubara2016binarized, rastegari2016xnor} propose to constrain both weights and activations to binary values (i.e., $+1$ and $-1$), where the multiplication-accumulations can be replaced by purely $\rm xnor$ and $\rm popcount$ operations, which are in general much faster. 
However, BNNs still suffer from significant accuracy 
decrease compared with the full precision counterparts. 
To narrow this accuracy gap, 
ternary networks \cite{li2016ternary,zhu2016trained} and even higher bit fixed-point quantization \cite{zhou2016dorefa, zhou2017incremental} methods are proposed.

In general, quantization approaches target at tackling two main problems. On one hand, some works target at designing more accurate quantizer to minimize information loss. For the uniform quantizer, works in \cite{choi2018pact,jung2019learning} explicitly parameterize and optimize the upper and/or lower bound of the activation and weights. To reduce the quantization error, non-uniform approaches \cite{park2017weighted, zhang2018lq} are proposed to better approximate the data distribution. 
In particular, LQ-Net \cite{zhang2018lq} proposes to jointly optimize the quantizer and the network parameters. 
On the other hand, because of the non-differentiable quantizer, some literature focuses on relaxing the discrete optimization problem. A typical approach is to train with regularization \cite{hou2018loss, zhou2018explicit, bai2019proxquant, ajanthan2019proximal, sakr2018true, choi2018learning}, where the optimization problem becomes continuous while gradually adjusting the data distribution towards quantized values.
Apart from the two challenges, with the popularization of %AutoML
neural architecture search (NAS),
Wang \etal \cite{wang2019haq} further propose to employ reinforcement learning to automatically determine the bit-width of each layer without human heuristics.

\noindent\textbf{Knowledge distillation.}
Knowledge distillation (KD) is a general approach for model compression, where a powerful wide/deep teacher distills knowledge to a narrow/shallow student to improve its performance~\cite{hinton2015distilling, romero2014fitnets}. 
% Moreover, some literature~\citep{Zhang_2018_CVPR, Park_2019_CVPR, GraphKD, Lee_2018_ECCV} designs advanced distillation strategies in order to let the student learn better from the teacher's rich knowledge.
% Due to the effectiveness of KD, it has also been widely used in many computer vision tasks. For example, Zhang~\etal~\citep{zhang2016real} propose to transfer the knowledge learned with optical flow CNN to improve the action recognition performance. 
% And several works propose to learn efficient object detection~\citep{chen2017learning, wei2018quantization} and semantic segmentation~\citep{he2019knowledge} with distillation. 
In terms of the definition of knowledge to be distilled from the teacher, existing models typically use teacher's class probabilities~\cite{hinton2015distilling} and/or intermediate features~\cite{romero2014fitnets, zagoruyko2016paying}. KD has been widely used in many computer vision tasks \cite{wei2018quantization, he2019knowledge}.
Moreover, there are some works \cite{zhuang2018towards, mishra2018apprentice, polino2018model} study the combination of KD and quantization, where the full-precision model provides hints to guide the low-precision model training and significantly improves the performance of the low-precision networks. Different from the previous literature, we propose a self-distillation strategy to improve our SP-Net training. Specifically, only the full-precision switch sees the ground-truth while the low-precision switches are learnt by distilling from the full-precision teacher. 
\section{Preliminaries}
\label{sec:preliminaries}

%%%%%%%%% BODY TEXT
\begin{figure*}
\begin{center}
   \includegraphics[width=1.0\linewidth]{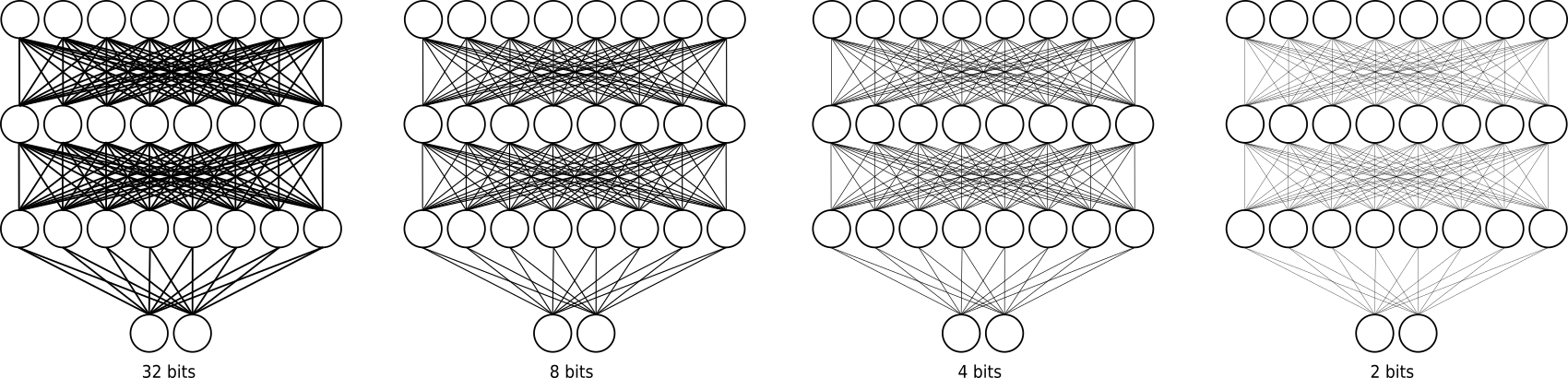}
\end{center}
   \caption{Overview of the proposed approach. In SP-Nets multiple precision switches share a common architecture, making it capable of adjusting the precision of their representations on the fly, granting devices and end-users real-time control over the network performance (thinner connections indicate reduced bitwidth).}
\label{fig:short}
\end{figure*}

The optimization problem of traditional Quantized Neural Networks (QNNs) aims at minimizing an objective function $\mL$ given a set of trainable weights $W$, which take values from $C_w$, a predefined discrete set typically referred as \textit{codebook}. A common QNN training procedure involves storing a real-valued version of $W$. During inference, the real $W$ is quantized using a predetermined pointwise quantization function $\rm{Quant_w}:\mathbb{R}\mapsto C_w$. The weights are updated during the optimization process by estimating the gradients w.r.t. the real-valued copy. Additionally, internal network activations can be optionally quantized with their own codebook $C_a$ by $\rm{Quant_a}:\mathbb{R}\mapsto C_a$. Given independent bitwidths $bits_w$ and $bits_a$ for the network weights and activations, $|C_w|=2^{bits_w}$ and $|C_a|=2^{bits_a}$. Finally, a quantized convolutional layer with $K$ filters is computed as follows:
\begin{equation}
\mathbf{O}_k=\sum_{i=1}^{I}\rm{Quant_a}(\mathbf{A}_{i,:,:})\ast \rm{Quant_w}(\mathbf{F}_{k,i,:,:}),
\label{eq:quantize_convolution}
\end{equation}
where $\mathbf F_k \in \mathbb{R}^{I\times h_{f}\times w_{f}} \subset W$ is the $k$-th convolutional filter. $I$, $h_{f}$ and $w_{f}$ denote the number of input channels, height and width of the filters, respectively. $\mathbf A\in \mathbb{R}^{I \times h_{in} \times w_{in}}$ and $\mathbf{O}_k \in \mathbb{R}^{h_{out} \times w_{out}}$ denote the input activations and output pre-activations of the filter, where $h_{in}$, $w_{in}$, $h_{out}$, $w_{out}$ represent the height and width of the input and output feature maps, respectively.

In the context of QNNs, multiple quantizers and gradient estimators have been proposed in the literature \cite{bengio2013estimating, liu2018bi, zhou2016dorefa, cai2017deep, miyashita2016convolutional}. For our SP-Net, we stick to common ones, which will be described in Sec. \ref{subsec:STE_and_quantizers}. Three non-linear quantizers for the activations are implemented. For the weights, the $\rm Tanh$-based quantizer is used, unless specified otherwise. The motivation to use each quantizer will be explained in each corresponding subsection.

\subsection{Straight-Through Estimator (STE) and Base Quantizer}
\label{subsec:STE_and_quantizers}
The STE proposed in \cite{bengio2013estimating} allows to estimate gradients through the non-differentiable functions $\rm round$ and $\rm sign$, making the $k$-bit quantizers employed compatible with the backpropagation algorithm. The STE operates by passing the output gradients unaltered to the input, it is equivalent to the derivative of an $\rm identity$ mapping on the inputs. Commonly, the gradients for inputs outside of the range $[-1,1]$ are suppressed and the derivative of the quantization function becomes equivalent to the derivative of the $\rm hard tanh$ function. 

The $k$-bit quantizers described in the next sections share the same base quantization function, $Q(\cdot)$:
\begin{equation}
Q(x)=\left\{\begin{matrix}
{\rm sign}(x) & \rm{if}\  k=1 \\ 
\frac{1}{2^k-1}{\rm round}((2^k-1)x) &{\rm if}\  k>1
\end{matrix}\right..
\label{eq:k_bit_quantizer}
\end{equation}
During backpropagation, we use the STE:
\begin{equation}
\frac{d\mL}{dx}\approx \frac{d\mL}{dQ} \mathbbm{1}_{|x|\leq 1}.
\label{eq:STE}
\end{equation}

\subsubsection{Tanh-Based Quantizer}
\label{sec:tanh_quantizer}
\cite{zhou2016dorefa, cai2017deep} proposed to use different quantizers for weights and activations. Weight quantizers approximate the hyperbolic tangent function ($\rm tanh$), constraining the weights to $[-1,1]$. However, $\rm tanh$ is associated with the vanishing gradient problem, thus, it is attractive for activations quantizers to approximate the popular $\rm ReLU$ activation function as described in the next section, constraining the activations to the positive range. 

The $\rm tanh$ is first used to project the input to range $[-1,1]$ in order to reduce the impact of large values. The quantizer is defined as follows:
\begin{equation}
{\rm TanhQuant}(x)=2\cdot Q(\frac{{\rm tanh}(x)}{2\cdot {\rm max}(|{\rm tanh}(x)|)}+\frac{1}{2})-1,
\label{eq:tanh_quantizer}
\end{equation}
where $Q(\cdot)$ hereafter is defined in Eq. (\ref{eq:k_bit_quantizer}).

It is appealing to train a network with activations switchable down to 1-bit with permissible values $\{-1,1\}$ since bit operations can be performed given that weights are 1-bit as well. Therefore, when training these type of networks, $\rm TanhQuant$ quantizer is used on both weights and activations to allow the activations of the intermediate switches to lie in the range $[-1,1]$. For this same reason, the layers were re-ordered as described in \cite{rastegari2016xnor} (typical layers ordering is {\rm QuantConv}$\rightarrow${\rm BN} $\rightarrow${\rm ReLU}$\rightarrow${\rm QuantConv}, while layers re-ordered are {\rm QuantConv}$\rightarrow${\rm ReLU}$\rightarrow${\rm BN}$\rightarrow${\rm QuantConv}).

\subsubsection{ReLU-Based Quantizer}
\label{sec:relu_quantizer}
As mentioned in the previous section, \cite{zhou2016dorefa, cai2017deep} employ a $\rm ReLU$ approximation quantizer for the activations with no layer re-ordering. The method proposed by \cite{zhou2016dorefa} will be employed here, which consists of simply clipping the activations followed by the base quantizer:
\begin{equation}
{\rm ReLUQuant}(x)=Q({\rm clip}(x,0,1)).
\label{eq:relu_quantizer}
\end{equation}

In particular, \cite{cai2017deep} proposed both uniform and non-uniform spacing between the codebook elements. They first clipped the activations to the range $[0,v]$, where $v$ is some predetermined value, and the codebook for both cases, uniform and non-uniform, is obtained from the network internal statistics. In our tests with $\rm ReLUQuant$ quantization, we simply constrain the activations to the range $[0,1]$ with uniform quantization. In the logaritmic quantizer described in the next section, non-uniform quantization is used.

\subsubsection{Logarithmic Quantizer}
\label{sec:log_quantizer}
In full-precision neural networks, the weights and activations have non-uniform distributions \cite{miyashita2016convolutional}. Taking advantage of that fact, the authors used logarithmic representations, both in weights and activations, achieving higher classification accuracy at the same resolution than uniform quantization schemes at an expense of higher implementation and computation complexity. In the original paper, their $\rm LogQuant$ layer contains a global Full Scale Range (FSR) parameter which is reported in the paper. Additionally, each layer has its own FSR. In our SP-Net, we use a variation of their $\rm LogQuant$ layer in order to avoid the FSR parameter. Our modified $\rm LogQuant$ quantizer is defined as follows:
\begin{equation}
{\rm LogQuant}(x)=\left\{\begin{matrix}
0 & {\rm if}\  x=0 \\ 
{2^{\widehat x}} \cdot x & {\rm otherwise}
\end{matrix}\right.,
\label{eq:log_quantizer}
\end{equation}
where
\begin{equation}
\hat{x}={\rm rescale}(Q({\rm normalize}(|{\rm log}(x)|))),
\end{equation}
\begin{equation}
{\rm normalize}(x)=(x- {\rm min}(x))/({\rm max}(x)-{\rm min}(x)),
\end{equation}
\begin{equation}
{\rm rescale}(x)=x\cdot ({\rm max}(x)-{\rm min}(x))+{\rm min}(x),
\end{equation}
\begin{equation}
s(x)={\rm sign}(x).
\end{equation}

Similarly to Sec. \ref{sec:tanh_quantizer}, two-sided logarithmic quantization (positive and negative) can be used in order to have activations switchable down to 1-bit. Depending on the choice of activations quantization (one-sided or two-sided), layer re-ordering should be taken into account.
In our experiments, logarithm base-2 was used, however, base-$\sqrt{2}$ could provide higher accuracy. 
\section{Switchable Precision Neural Networks}
\label{sec:quantizable_neural_network}

%%%%%%%%% BODY TEXT
\textit{SP-Nets} generalize QNNs, where the learnable weights $W$ are optimized for multiple codebooks $C_w[n]$ and $C_a[n]$ of variable cardinality. %\bohan{$n=1,...,N$ } \luis{I'm going to remove this because if gives the impression that n is the cardinality}. 
The permissible values of the $N$ codebooks are determined by the choice of quantizers $\rm Quant_w$ and $\rm Quant_a$, while their cardinality is determined by the bitwidths $bits_w[n]$ and $bits_a[n]$.

DNNs at their current state are not naturally precision switchable as empirically demonstrated in Sec. \ref{textbf:non-switchable_batchnorm}. Therefore, we appeal to S-BN, a technique used to train slimmable neural networks, described in detail in Sec.  \ref{subsec:switchable_batchnorm}.
Then in Sec. \ref{subsec:slimmable/quantizable}, we extend our SP-Net to mixed slimmable SP-Net. 
%\luis{I didn't write self-distillation here because it is a different section}

\subsection{Switchable Batch Normalization (S-BN)}
\label{subsec:switchable_batchnorm}
DNNs at their current state are not naturally slimmable \cite{yu2019universally} nor of switchable precision due to the inconsistent behavior of batch normalization layers during training and inference. During training, BN layers utilize the current batch $x_b$ mean $E_b[x_b]$ and variance $Var_b[x_b]$ to perform intermediate feature maps normalization while accumulating a running mean $u$ and running variance $\sigma^2$, which is used as replacement during test.

In a naive SP-Net, the mini-batch local statistics of each quantization switch are different, however, all the switches will contribute to the accumulated global mean $u$ and variance $\sigma^2$, thus, enabling the network to train properly, but resulting in inaccurate test inference. 

S-BN layers equip regular BN layers with private $u[n]$ and $\sigma^2[n]$ for each of the $N$ switches. The overhead of the additional parameters is negligible since BN parameters account for an insignificant portion of the total amount. It is also negligible in terms of run-time complexity since they involve no additional operations.

Additionally, BN layers count with two learnable parameters $\beta$ and $\gamma$ used to provide the layer with the capacity of performing a linear mapping. Unlike $u$ and $\sigma^2$, $\beta$ and $\gamma$ can be updated by all the switches. Providing with private versions of them is not as crucial, since they allow the network to learn, but it yields an additional accuracy boost \cite{yu2019universally}. Furthermore, they generate no additional overhead since they can be merged with $u[n]$ and $\sigma^2[n]$ after training.
%\bohan{Not crucial is not accurate since you never tested it.} \luis{The original paper has numbers, that's where I saw it}

In Algorithm \ref{algo:quantizable_net_training}, we illustrate the use of S-BN in one SP-Net training iteration.

\begin{algorithm}[htp!]
\SetAlgoLined
\textbf{Require:} SP-Net $M$. Lists of weights and activations bitwidths $bits\_w$, $bits\_a$. \\
Get mini-batch data $x$ and ground-truth $y$;\\
\For{bit\_w in bits\_w}{
    \For{bit\_a in bits\_a}{
        Switch BN parameters in $M$ for the current bitwidths;\\
        Set $bit\_w$, $bit\_a$ in $M$;\\
        Forward pass using input $x$;\\
        Compute loss w.r.t. to $y$;\\
        Compute gradients using STE;
    }
}
Update weights using accumulated gradients;
\caption{One SP-Net iteration using S-BN}
\label{algo:quantizable_net_training}
\end{algorithm}

\subsection{Slimmable SP-Net}
\label{subsec:slimmable/quantizable}

Network slimming \cite{yu2018slimmable} relies on S-BN in order to allow each layer of the network to operate at different widths. Given the current width multiplier $width\in[0,1]$, a slimmable convolutional layer with $K$ filters computes only the first $\lceil K*width \rceil$ ones.

Network slimming and quantization are complementary techniques and effortlessly work along without technical complications.
Therefore, we can train a single shared network with switchable width and precision to increase the flexibility.
A single slimmable/SP network can be trained in the cloud and distribute particular switches to different deployment systems based on their specific hardware capabilities, where they can be further slimmed and quantized instantaneously on demand. In Sec. \ref{textbf:slimmable_quantizable_testing}, we provide a comparison of a slimmable SP-Net with the corresponding individually trained switches.

Although the benefits of a SP-Net in terms of power and speed improvements are evident by performing dot-product operations on quantized vectors, the benefits on memory footprint may not be so apparent, given that the full precision weights must be stored on non-volatile memory at all time, regardless of the active quantization switch. However, during operation, a network clone is stored in the RAM of the processor in order to provide quick access, therefore weight quantization only takes place once every time a new switch is requested and the quantized clone is kept in volatile memory. By quantizing activations, additional RAM savings can be obtained.

\section{Self-Distillation}
\label{sec:self-distillation}
Knowledge distillation is a common strategy used to provide a stronger training signal, typically from a large network to smaller one in a teacher-student scheme. In a similar fashion to knowledge distillation, by simultaneously optimizing multiple quantized switches, implicitly the high precision switches are providing guidance to the noisier low precision updates. However, this behavior is not explicitly encouraged. Therefore, in this section we present a complementary distillation mechanism, denominated \textit{self-distillation}, where only the full-precision switch sees the ground-truth, while the low bitwidth switches try to match the internal representation as well as the output distribution of the full-precision switch. The strategy formulated allows the full-precision switch to guide the optimization process with a significant amount of information flow across all the switches.

In US-Net \cite{yu2019universally} gradients are prevented from flowing from the sub-networks to the largest width switch. Formally, to mimic the outputs, similarly to US-Net, we use the Kullback–Leibler divergence as distance measure on the output distributions $p_r$ and $p_q$ of the full-precision switch and the quantized active switch. Let $SG(\cdot)$ denote the stop-gradient function, the output mimic loss is:
\begin{equation}
\mL_{out}=D_{KL}(SG(p_r)\lVert p_q).
\end{equation}

Additionally, to create the guidance signal, \cite{zhuang2018towards} proposes a hint-based training strategy by comparing the intermediate feature maps between the full-precision teacher and the low-precision student. Similarly, let $f_r$ and $f_q$ denote the internal feature maps (\ie, pre-activations) of the full-precision switch and the quantized active switch, the internal representations guidance loss becomes:
\begin{equation}
\mL_{f}={\left\lVert f_r-f_q\right\rVert}^2_2.
\end{equation}
%In our experiments we tried both stopping the gradients flowing to $f_r$ and quantizing $f_r$. 
%However, both strategies incurred in lower accuracy in this case. 
And in this case, we do not stop the gradients flowing to $f_r$ and quantize $f_r$. 

%perform a direct comparison between features maps.
Finally, the loss for the quantized switches is their combination, while for the full precision switch is simply the regular cross-entropy classification loss:
\begin{equation}
\mL_q=\alpha_1\mL_{out}+\alpha_2\mL_{f},
\end{equation}
\begin{equation}
\mL_r=\mL_{cross-entropy}(p_r, y),
\end{equation}
where $y$ denotes the ground-truth labels, and $\alpha_1$ and $\alpha_2$ are the scaling coefficients to balance $\mL_{out}$ and $\mL_{f}$, respectively.

\section{Experiments}
In this section, we test 2 different achitectures, ResNet \cite{he2016deep} and MobileNet \cite{howard2017mobilenets}, for the task of image classification on the ImageNet (ILSVRC-2012) \cite{russakovsky2015imagenet} and Tiny ImageNet datasets with 3 different quantizers: $\rm Tanh$-based quantization \cite{zhou2016dorefa}, $\rm ReLU$-based quantization \cite{zhou2016dorefa, cai2017deep} and non-uniform Logarithmic quantization \cite{miyashita2016convolutional}. Experiments on ImageNet have been included in the supplementary material.

We implement our SP-Nets using Pytorch. For our ImageNet experiments, we use a regular (\ie, 1 single switch) full-precision pre-trained model, we then replicate the BN parameters across all the S-BN switches and fine-tune the SP-Net model. We use the standard pre-processing and augmentation as reported by \cite{he2016deep}. For training with the pre-trained full-precision model, we use Adam \cite{kingma2014adam} optimizer with an initial learning rate of $1e^{-4}$ and decrease it by a factor of 10 at epochs 15 and 20 with a total of 25 epochs. For our Tiny ImageNet experiments, all the networks are learned from scratch using SGD optimizer with initial learning rate of 0.1 and decreased by a factor of 10 every 30 epochs during 100 epochs. As in common practice, the first and last layers are not quantized \cite{rastegari2016xnor}.
%\bohan{where are $\alpha_1$ and $\alpha_2$?} \luis{In the self-distillation subsection}

\noindent\textbf{Implementation Note}. The activations of the full-precision switch, although are not quantized, in some cases they must still be clipped to the same range of the quantized switches, as it can lead the switch to diverge.

\subsection{Evaluation on ImageNet}
\paragraph{Evaluation on ResNet-18}
In Table \ref{tab:Resnet_18-Imagenet_edge}, we compare the Top-1 accuracy for the edge bitwidth cases in both weights and activations for a ResNet-18 architecture. Our SP-net achieved very close accuracy to the independently trained QNNs with the full-precision switch being the most affected, while providing increased efficiency-accuracy flexibility.

\begin{table}[hbt!]
	\small
	\begin{center}
	\caption{Top-1 accuracy (\%) for ResNet-18 on ImageNet.}
	\scalebox{0.9}{
		\begin{tabular}{c c|c c}
			\hline
			Weight&Activation&SP-Net&Independent\\
			\hline
			2 & 2 & 62.4 & \textbf{62.6} \\
			2 & 32 & 65.3 & \textbf{65.8} \\
			32 & 2 & 65.1 & \textbf{65.3} \\
			32 & 32 & 68.1 & \textbf{69.5} \\
			\hline
		\end{tabular}}
		\label{tab:Resnet_18-Imagenet_edge}
     \end{center}
     \vspace{-1.0em}
\end{table}

% We additionally investigate the effect of the bitwith gap in SP-Nets. In Table \ref{tab:Resnet_18-Imagenet_short_gap}, we present results with a shorter gap, using bitwidths of a more practical scenario. Consistently with the edge cases, our SP-net achieved very similar results to independently trained QNNs, and the bitwidth gap does not have an impact in the accuracy.

% \begin{table}[hbt!]
% 	\small
% 	\begin{center}
% 	\caption{Top-1 accuracy (\%) with a short gap between switches for Resnet-18 on ImageNet.}
% 	\scalebox{0.9}{
% 		\begin{tabular}{c c|c c}
% 			\hline
% 			Weight&Activation&SP-Net&Independent\\
% 			\hline
% 			2 & 2 & 62.5 & \textbf{62.6} \\
% 			2 & 4 & 64.6 & \textbf{64.9} \\
% 			3 & 2 & 63.9 & \textbf{64.0} \\
% 			3 & 4 & 66.3 & \textbf{66.5} \\
% 			\hline
% 		\end{tabular}}
% 	 \label{tab:Resnet_18-Imagenet_short_gap}
%      \end{center}
%      \vspace{-1.0em}
% \end{table}

\noindent\textbf{Evaluation on MobileNet.}
The MobileNet architecture makes use of the depthwise separable and pointwise convolutions without residual connections drastically reducing the number of parameters without a major decrease in accuracy. However, it is very sensitive to quantization, specially on the activations. \cite{sheng2018quantization} re-designs the MobileNet architecture to be more quantization friendly. For our MobileNet SP-Net, we follow their architecture. The re-design involves simple layers replacement and re-ordering. In Table \ref{tab:Mobilenet-Imagenet}, we report the results of different bitwidths for weights and activations using our SP-Net MobileNet and independent networks with the modified architecture. As can be observed, the impact of switchable precisions on MobileNet is higher than on ResNet-18.
\begin{table}[hbt!]
	\small
	\begin{center}
	\caption{Top-1 accuracy (\%) for MobileNet on ImageNet.}
	\scalebox{0.9}{
		\begin{tabular}{c c|c c}
			\hline
			Weight&Activation&SP-Net&Independent\\
			\hline
			8 & 8 & 59.5 & \textbf{63.5} \\
			8 & 32 & 66.0 & \textbf{69.0} \\
			32 & 8 & 59.6 & \textbf{63.6} \\
			32 & 32 & 66.1 & \textbf{69.7} \\
			\hline
		\end{tabular}}
		\label{tab:Mobilenet-Imagenet}
     \end{center}
     \vspace{-1.0em}
\end{table}

\subsection{Evaluation on Tiny ImageNet}
The Tiny ImageNet dataset is a downsampled ImageNet version ($64\times64$) with 100K images for training and 10K for validation, spread across 200 classes.
For the experiments on Tiny ImageNet, we used the same ResNet-18 architecture as that for ImageNet except for the first layer, whose filter size is $3\times3\times64$ with $stride = 1$ and $padding = 1$.

\begin{table}[ht]
	\small
	\begin{center}
	\caption{Top-1 accuracy (\%) for a 1-bit SP-Net ResNet-18 on Tiny ImageNet.}
	\scalebox{0.9}{
		\begin{tabular}{c c|c c}
			\hline
			Weight&Activation&SP-Net&Independent\\
			\hline
			1 & 1 & \textbf{43.5} & 40.0 \\
			1 & 3 & \textbf{48.8} & 46.3 \\
			1 & 8 & \textbf{48.8} & 47.0 \\
			1 & 32 & 49.0 & \textbf{49.1} \\
			\hline
		\end{tabular}}
		\label{tab:1bit_network}
     \end{center}
     \vspace{-1.0em}
\end{table}

\begin{figure}[t]
\begin{center}
   \includegraphics[width=0.9\linewidth]{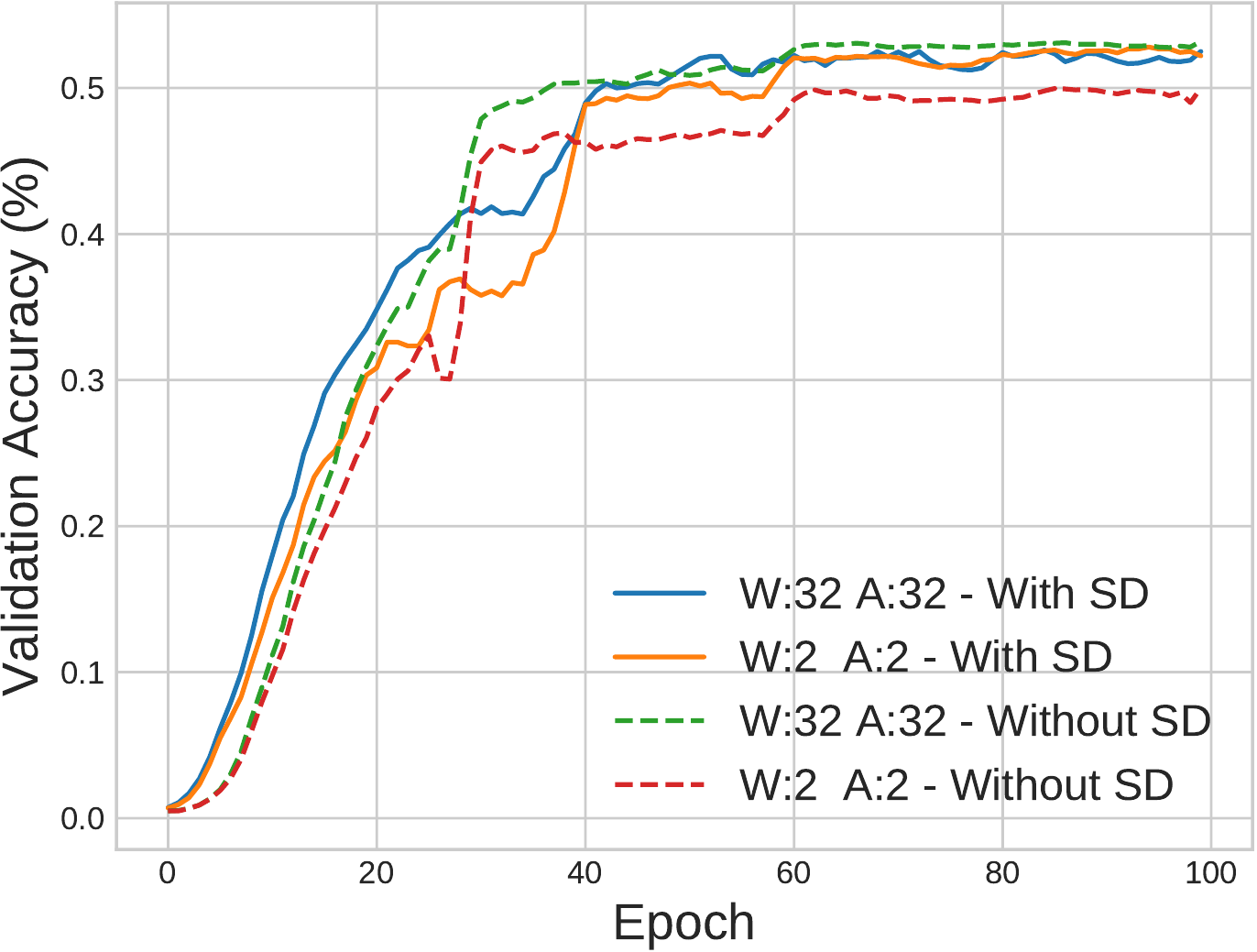}
\end{center}
   \caption{Top-1 accuracy for SP-Net ResNet-18 with and without self-distillation (SD) on Tiny ImageNet}
\label{fig:self_distillation}
\end{figure}

\noindent\textbf{Comparison of Different Quantizers.}
\begin{table*}[hbt!]
	\small
	\begin{center}
	\caption{Comparison of different quantizers on a SP-Net for ResNet-18 on Tiny ImageNet. The networks were slimmed with a factor of $0.25$ to magnify the effect of the quantizers.
	$^1$ $\rm ReLU$-based: $\rm TanhQuant$ quantizer on weights, $\rm ReLUQuant$ on activations.
	$^2$ $\rm Tanh$-based: $\rm TanhQuant$ quantizer on both weights and activations \textit{with layer re-ordering}.
	$^3$Logarithmic 1: $\rm TanhQuant$ quantizer on weights and $\rm LogQuant$ quantizer on activations.
	$^4$Logarithmic 2: $\rm LogQuant$ quantizer on both on weights and activations.}
	\scalebox{0.9}{
		\begin{tabular}{c|c c|c c c c c}
		    \hline
		    Width&\multicolumn{2}{c|}{Bitwidth}&\multicolumn{4}{c}{Quantizer Accuracy(\%)} \\
		    \hline
			&Weight&Activation&ReLU-based$^1$&Tanh-based$^2$&Logarithmic $1^3$& Logarithmic $2^4$\\
			\hline
			0.25 & 2 & 2 & \textbf{32.5} & 29.9 & 29.1 & 28.9\\
			0.25 & 2 & 4 & \textbf{34.5} & 32.7 & 34.0 & 33.9\\
			0.25 & 2 & 8 & \textbf{34.8} & 32.8 & 34.1 & 34.1\\
			0.25 & 2 & 32 & 34.3 & 32.8 & 34.4 & \textbf{35.5}\\
			0.25 & 4 & 2 & \textbf{34.1} & 33.7 & 33.3 & 32.5\\
			0.25 & 4 & 4 & \textbf{37.4} & 36.8 & 37.3 & 36.7\\
			0.25 & 4 & 8 & 37.4 & 36.6 & \textbf{37.5} & 37.0\\
			0.25 & 4 & 32 & 37.5 & 36.6 & 37.6 & \textbf{37.7}\\
			0.25 & 8 & 2 & \textbf{34.4} & 33.0 & 33.9 & 32.6\\
			0.25 & 8 & 4 & \textbf{37.3} & 36.5 & 37.2 & 36.4\\
			0.25 & 8 & 8 & 37.3 & 36.9 & \textbf{37.4} & 36.9\\
			0.25 & 8 & 32 & 37.5 & 37.0 & 37.6 & \textbf{37.7}\\
			0.25 & 32 & 2 & \textbf{36.3} & 33.6 & 33.5 & 32.8\\
			0.25 & 32 & 4 & \textbf{37.3} & 36.5 & 37.2 & 36.8\\
			0.25 & 32 & 8 & 37.2 & 37.0 & \textbf{37.4} & 36.9\\
			0.25 & 32 & 32 & \textbf{37.7} & 37.0 & 37.5 & 37.5\\
			\hline
		\end{tabular}}
		\label{tab:quantizers_comparisson}
     \end{center}
     \vspace{-1.0em}
\end{table*}

As explained in Sec. \ref{subsec:STE_and_quantizers}, different quantizers can be used in different scenarios. $\rm ReLUQuant$ quantizer is used in activations in general. When it is desirable to have activations quatizable to 1-bit, $\rm TanhQuant$ quantizer is used. $\rm LogQuant$ quantizer is chosen when higher accuracy is desired with the same bitwidth, at an expense of higher complexity.

In Table \ref{tab:quantizers_comparisson}, 4 quantizer configurations were tested for SP-Nets with multiple weight and activation switches
. In our results, $\rm Tanh$-based configuration obtained the lowest accuracy across all the switches and $\rm ReLU$-based configuration frequently obtained the best results. Logarithmic based configurations ocasionally performed better, particularly at higher activations bitwidths. Logarithmic based configurations were expected to produce the best results, however our FSR-free modified version and logarithm base choice may have harmed their performance. All the networks were slimmed by a factor of $0.25$ to magnify the effect of the quantizers.

\noindent\textbf{1-bit SP-Net.}
We trained a network with a single bit per weight and switchable precision activations. Our network spans from the popular BinaryConnect, where dot products are computed using only summation, to Binarized Neural Networks (BNNs), where activations are quantized to 1-bit and dot products are computed using $\rm xnor$ and $\rm popcount$ operations. 

Training a network with precision switchable down to 1-bit per activation is particularly challenging since it causes a significant decrease in performance across all the bitwidths when using a $\rm ReLU$-based quantization. Therefore, networks with activations switchable to 1-bit are trained with $\rm Tanh$-based quantization with layer re-ordering. Weights are trained by simply using the $\rm sign$ function with STE.

The results in Table \ref{tab:1bit_network} show that our 1-bit SP-Net surpasses the independently trained counterparts by a large margin when the activations are in extremely low-precision.

\noindent\textbf{Slimmable SP-Net.} Our results for slimmable SP-Net in Table \ref{tab:slimmable/quantizable} demonstrate the flexibility of our network, however they show no clear accuracy advantage over independently trained networks. We tested widths of $0.25$, $0.5$ and $1.0$, with slimmable SP-Net and independently trained networks constantly outperforming each other. However, it is worth noting that our slimmable SP-Net network can simultaneously switch width and bitwidth on demand.

\label{textbf:slimmable_quantizable_testing}
\begin{table}[hbt!]
	\small
	\begin{center}
	\caption{Top-1 accuracy (\%) for a slimmable SP-Net ResNet-18 on Tiny ImageNet.}
	\scalebox{0.9}{
		\begin{tabular}{c|c c|c c}
			\hline
			Width&Weight&Activation&Slim/SP-Net&Independent\\
			\hline
			0.25 & 2 & 2 & \textbf{30.3} & 26.5 \\
			0.25 & 2 & 32 & \textbf{36.5} & 34.2 \\
			0.25 & 32 & 2 & \textbf{35.6} & 32.8\\
			0.25 & 32 & 32 & \textbf{39.7} & 37.8 \\
			0.5 & 2 & 2 & 40.9 & \textbf{41.3} \\
			0.5 & 2 & 32 & 44.8 & \textbf{49.7} \\
			0.5 & 32 & 2 & \textbf{46.1} & 45.8 \\
			0.5 & 32 & 32 & 49.0 & \textbf{52.8} \\
			1.0 & 2 & 2 & \textbf{47.8} & 47.4 \\
			1.0 & 2 & 32 & 50.5 & \textbf{54.4} \\
			1.0 & 32 & 2 & \textbf{51.8} & 50.7 \\
			1.0 & 32 & 32 & 53.0 & \textbf{56.4} \\
			\hline
		\end{tabular}}
		\label{tab:slimmable/quantizable}
     \end{center}
     \vspace{-1.0em}
\end{table}

\noindent\textbf{Self-Distillation.} 
Our self-distillation method in Sec. \ref{sec:self-distillation} proved to be very effective, by improving the accuracy across all the switches, as can be observed in Table \ref{tab:self-distillation}. Additionally it can be observed that the validation performance of intermediate switches is superior to the performance of the full-precision one, however we credit this to over-fitting, since the training accuracy is higher for the full precision switches. We also compared the performance of the distillation losses, confirming our feature maps distillation loss $\mL_{f}$ is indeed benefiting the overall performance of the network. The value of $\mL_{f}$ is expected to be several orders of magnitude larger than $\mL_{out}$, since it is dictated by the number of layers and the size of the intermediate feature maps. We set the hyper-parameters $\alpha_1=1$ and $\alpha_2=1e^{-7}$ to bring $\mL_{f}$ to the same order of magnitude with $\mL_{out}$. In Figure \ref{fig:self_distillation}, we plot the accuracy curves during training of our SP-Net with and without self-distillation.

\begin{table}[ht]
	\small
	\begin{center}
	\caption{Top-1 accuracy (\%) for a SP-Net network with and without self-distillation for ResNet-18 on Tiny ImageNet.}
	\scalebox{0.9}{
		\begin{tabular}{c c|c c c}
			\hline
			Weight&Activation&Regular&$L_{out}$&$L_{out}+L_{f}$\\
			\hline
			2 & 2 & 50.1 & 53.0 & \textbf{53.3}\\
			2 & 3 & 50.5 & 53.8 & \textbf{53.9}\\
			2 & 32 & 51.2 & \textbf{54.2} & 53.8\\
			3 & 2 & 50.8 & 53.4 & \textbf{53.9}\\
			3 & 3 & 51.5 & 54.0 & \textbf{54.4}\\
			3 & 32 & 52.3 & 54.1 & \textbf{54.5}\\
			32 & 2 & 51.4 & 52.6 & \textbf{53.3}\\
			32 & 3 & 52.1 & 53.4 & \textbf{53.9}\\
			32 & 32 & 52.9 & \textbf{52.9} & 52.8\\
			\hline
		\end{tabular}}
		\label{tab:self-distillation}
     \end{center}
     \vspace{-1.0em}
\end{table}

\subsection{Ablation Studies}
\label{subsec:ablation_studies}

\begin{figure}[t]
\begin{center}
   \includegraphics[width=0.9\linewidth]{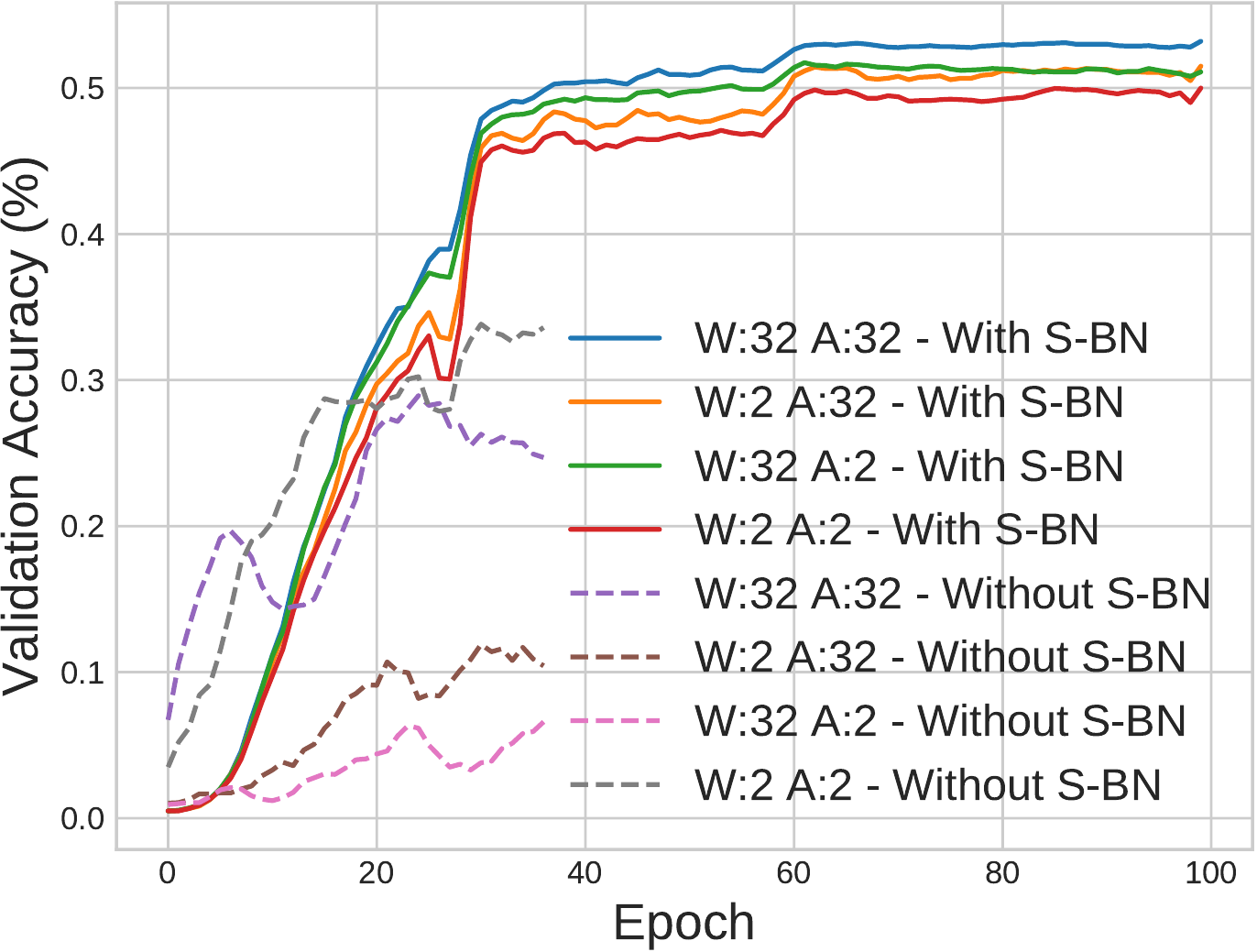}
\end{center}
   \caption{Top-1 accuracy for SP-Net ResNet-18 with and without switchable batch normalization (S-BN) on Tiny ImageNet.}
\label{fig:non-swtichable_batchnorm}
\end{figure}

\noindent\textbf{Non-Switchable Batch Normalization.}
We replaced our S-BN layers with standard BN ones to verify and demonstrate that privatizing all BN layers for each switch is essential to successfully perform inference. Figure \ref{fig:non-swtichable_batchnorm} shows the validation plots for a SP-Net network with 4 switches with and without S-BN. The switches of the network with S-BN learn at different rates, with the high precision ones learning faster and achieving higher accuracy than the low precision ones, while the accuracy of the switches without S-BN quickly stagnates and does not recover.
\label{textbf:non-switchable_batchnorm}

\noindent\textbf{Impact of Multiple Switches.} The impact of increasing the number of switches was investigated. In Table \ref{tab:multiple_switches}, we present the results of SP-Net with 4 and 16 switches slimmed with a factor of $0.5$. The first remark that we notice is that with increasing number of switches, the initial learning rate should be lowered. For the network with 16 switches, we set the initial learning rate to $0.03$. The accuracy obtained uses less switches is higher, we conjecture is due to the additional constraints, however it indicates an interesting direction of research.
\begin{table}[hbt!]
	\small
	\begin{center}
	\caption{Top-1 accuracy (\%) of SP-Nets with 4 and 16 switches slimmed with a factor of $0.5$.}
	\scalebox{0.95}{
		\begin{tabular}{|c|c|c c c c | c c c c|}
		    \cline{3-10}
		    \multicolumn{2}{c|}{}&\multicolumn{4}{c|}{Weights bitwidth}&\multicolumn{4}{c|}{Weights bitwidth} \\
		    \cline{3-10}
			\multicolumn{2}{c|}{}&2&4&8&32&2&4&8&32\\
			\hline
			\multirow{4}{*}{\rotatebox[origin=c]{90}{\shortstack[c]{Activations\\bitwidht}}} 
			& 2 & \textbf{41.3} & - & - & \textbf{43.8} & 37.7 & 41.3 & 41.4 & 41.4 \\
			& 4 & - & - & - & - & 38.6 & 42.3 & 42.9 & 42.4 \\
			& 8 & - & - & - & - & 38.4 & 42.5 & 42.7 & 42.5 \\
			& 32& \textbf{43.3} & - & - & \textbf{45.2} & 39.6 & 42.7 & 43.2 & 43.1 \\
			\hline
		\end{tabular}}
		\label{tab:multiple_switches}
     \end{center}
     \vspace{-1.0em}
\end{table}

\noindent\textbf{Switchable Precision in Weights vs Activations.}  \cite{zhu2019binary} found that QNNs activations are more sensitive to quantization than weights. Our observations on a SP-Net MobileNet confirm their findings. However, for our SP-Net ResNet-18, we observe that weights and activations are about equally sensitive. For example in Table \ref{tab:quantizers_comparisson}, by freezing the weights to $2$-bit and varying the bitwidths of activations in $\{2, 4, 8, 32\}$, we can observe that the accuracy gap is $1.8\%$ with top accuracy of $34.3\%$ for the $\rm ReLU$-based quantizer, and gap of $6.6\%$ with top accuracy of $35.5\%$ for Logarithmic 2 quantizer. Similarly, by freezing the activations to 2-bit and varying the bitwidths of weights, the gap for $\rm ReLU$-based quantizer is $3.8\%$ with top accuracy of $36.3\%$, and for Logarithmic 2 the gap is $3.9\%$ with top accuracy of $32.8\%$. 

Additionally, in Figure \ref{fig:non-swtichable_batchnorm}, we can observe that the switch ``W:2 A:32'' learns slower than ``W:32 A:2'', but towards the end of training, it catches up.

\section{Conclusion and Future Work}
In this paper we have proposed a DNN capable of operating at variable precisions on demand. With this approach, we grant devices and end-users real-time control over the performance of the DNNs powering inference algorithms. Our approach is lightweight and does not require altering the model, making it compatible with connectivity-free and limited memory devices. An additional virtue of our proposed network lies in ability to train a single network that can be distributed to different devices based on their capabilities. We have demonstrated the flexibility of our method with multiple quantization functions and a slimming complementary strategy. Moreover, we have proposed a training procedure to increase the accuracy of our network across the available precisions. Finally, we performed multiple ablation studies to analyse the performance of our approach in different scenarios.

Following the spirit of Universally Slimmable networks, a continuously quantizable SP-Net would be an interesting research direction as well as mixed precision SP-Net, where each layer is quantized independently on-demand. This method could be used to find the optimal layer-wise bitwidth for weights and activations.
\section{Acknowledgements}
This work was supported by the Australian Research Council through the ARC Centre of Excellence for Robotic Vision (project number CE1401000016).

{\small
\bibliographystyle{ieee_fullname}
\bibliography{egbib}
}

\end{document}